\documentclass[conference]{IEEEtran}
\IEEEoverridecommandlockouts
\usepackage{cite}
\usepackage{amsmath,amssymb,amsfonts}
\usepackage{algorithmic}
\usepackage{textcomp}
\usepackage{xcolor}

\usepackage[pdftex]{graphicx}
\usepackage{color}
\usepackage{multirow}
\usepackage{algorithm}
\usepackage{multibib}

\DeclareMathOperator*{\Dir}{Dir}

\DeclareMathOperator*{\Mult}{Mult}
\DeclareMathOperator*{\Cat}{Cat}

\def\BibTeX{{\rm B\kern-.05em{\sc i\kern-.025em b}\kern-.08em
    T\kern-.1667em\lower.7ex\hbox{E}\kern-.125emX}}
\begin{document}

\title{Symbol Emergence as Inter-personal Categorization with Head-to-head Latent Word
\thanks{This work was supported by JSPS KAKENHI Grant Number 21H04904.}
}

\author{
\IEEEauthorblockN{Kazuma Furukawa}
\IEEEauthorblockA{\textit{Graduate School of Information Science and Engineering} \\
\textit{Ritsumeikan University}\\
Kusatsu, Japan\\
furukawa.kazuma@em.ci.ritsumei.ac.jp}
\and
\IEEEauthorblockN{Akira Taniguchi}
\IEEEauthorblockA{\textit{College of Information Science and Engineering} \\
\textit{Ritsumeikan University}\\
Kusatsu, Japan\\
a.taniguchi@em.ci.ritsumei.ac.jp}
\and
\IEEEauthorblockN{Yoshinobu Hagiwara}
\IEEEauthorblockA{\textit{College of Information Science and Engineering} \\
\textit{Ritsumeikan University}\\
Kusatsu, Japan\\
yhagiwara@em.ci.ritsumei.ac.jp}
\and
\IEEEauthorblockN{Tadahiro Taniguchi}
\IEEEauthorblockA{\textit{College of Information Science and Engineering} \\
\textit{Ritsumeikan University}\\
Kusatsu, Japan\\
taniguchi@em.ci.ritsumei.ac.jp}
}

\maketitle

\begin{abstract}

In this study, we propose a head-to-head type (H2H-type) inter-personal multimodal Dirichlet mixture (Inter-MDM) by modifying the original Inter-MDM, which is a probabilistic generative model that represents the symbol emergence between two agents as multiagent multimodal categorization. A Metropolis--Hastings method-based naming game based on the Inter-MDM enables two agents to collaboratively perform multimodal categorization and share signs with a solid mathematical foundation of convergence. 
However, the conventional Inter-MDM presumes a tail-to-tail connection across a latent word variable, causing inflexibility of the further extension of Inter-MDM for modeling a more complex symbol emergence. 
Therefore, we propose herein a head-to-head type (H2H-type) Inter-MDM that treats a latent word variable as a child node of an internal variable of each agent in the same way as many prior studies of multimodal categorization. On the basis of the H2H-type Inter-MDM, we propose a naming game in the same way as the conventional Inter-MDM. The experimental results show that the H2H-type Inter-MDM yields almost the same performance as the conventional Inter-MDM from the viewpoint of multimodal categorization and sign sharing. 
\end{abstract}


\section{Introduction}
Understanding the fundamental evolutionary and developmental dynamics of symbol emergence and language communication is a crucial challenge in cognitive and developmental robotics. Symbol emergence is a phenomenon in which communication using semiotic signs that have meanings and contribute to the agents' adaptation to the environment emerges. Accordingly, many computational studies have been conducted in developmental and evolutionary systems including robotics~\cite{Steels95,Steels15,Steels05,Spranger12,Vogt02,Vogt05,Beule06,Bleys15,Matuszek18}.
Recently, many studies based on deep reinforcement learning have been conducted to model emergent communications as well~\cite{kim2021communication,Jiang,li2019ease,eccles2019biases,simoes2019multi,cowen2020emergent,lazaridou2020emergent}.
Among them, a language game is one popular approach for modeling the symbol emergence~\cite{Steels15}. 
Naming games that allow two or more agents to interact with each other, form object categories, and share names (i.e., signs) are a representative game among language games~\cite{Steels15}. 

Hagiwara et al. proposed a naming game based on a probabilistic generative model (PGM) and the Metropolis--Hastings method, which is a type of Markov Chain Monte Carlo algorithms~\cite{Hagiwara19}.
We call this naming game the Metropolis--Hastings method-based (MH-based) naming game.
The PGM enabling agents to perform multimodal categorization and the MH-based naming game is called an inter-personal multimodal Dirichlet mixture (Inter-MDM)~\cite{Furukawa20}. With the solid theoretical basis of the Metropolis--Hastings method, the naming game is guaranteed to perform sign sharing between agents possible in a probabilistic manner. 
The Inter-MDM is a model combining two PGMs for multimodal categorization by two agents into a single PGM. Therefore, the symbol emergence process is regarded as an inference process of the shared word variable from a Bayesian viewpoint.
This provides a theoretical insight on the symbol emergence. This view allows us to consider symbol emergence from the viewpoint of the generative view of cognition. Recently, predictive coding and the free-energy principle are often regarded as a general principle of cognition~\cite{ciria2021predictive,FRISTON2021573}.
The view based on the PGM provides us with a theoretical connection between symbol emergence and the free-energy principle. 

We refer to the conventional Inter-MDM herein as the Tail-to-Tail type Inter-MDM because it has a tail-to-tail latent word variable (see Figure~\ref{fig:t2t_gm}). 
The latent word is a random variable connecting the original two multimodal Dirichlet mixtures (MDMs) that represents the word that emerges through the naming game. 

However, the conventional Inter-MDM theoretically has the following two problems.
\begin{itemize}
    \item If the variable representing a sign must be a variable for a prior distribution of the internal variables of an agent, such a constraint about the design of variables and distributions for signs will become a bottleneck for further model extension (e.g., treating sentences and images as signs) because it is easier to treat sentences, images, and other complex observations as the output of a probabilistic distribution than a prior distribution.
    \item The manner of treating signs as a prior variable is inconsistent with that in the previous works of multimodal categorization and representation learning based on PGMs. Most studies treated signs (e.g., words and sentences) as a child node of an agent's internal variable. This difference prevented us from connecting findings prior to the works related to multimodal categorization with those of symbol emergence based on the Inter-MDM and its peripheral works.
\end{itemize}

One question, however, arises. Suppose that we change the latent word variable of the Inter-MDM from a prior variable to a posterior variable (i.e., from a parent node to a child node in a graphical model). Can the total Inter-MDM still reproduce the symbol emergence as multiagent multimodal categorization? Does it enable two agents to perform multimodal categorization and share signs?

To answer this question, we devised a modified version of the Inter-MDM, whose latent word is treated as a child node of an internal variable of an agent instead of a parent node. 
In other words, we changed the tail-to-tail connection across the latent word to a head-to-head connection. 
We call this modified Inter-MDM head-to-head type (H2H-type) as the Inter-MDM.

The contributions of this paper are
\begin{enumerate}
    \item We propose an H2H-type Inter-MDM and derive an MH-based naming game in the manner similar to that of the T2T-type Inter-MDM. 
    \item We conduct an experiment using synthetic data and show that the H2H-type Inter-MDM enables two agents to collaboratively perform multimodal categorization and share signs in the same way as the T2T-type Inter-MDM.
\end{enumerate}

The remainder of this paper is structured as follows:
Section 2 describes the preliminaries and introduces the T2T-type Inter-MDM and the SERKET framework; Section 3 presents the H2H-type Inter-MDM and the proposed MH-based naming game; Section 4 explains the experiment comparing the T2T- and H2H-type Inter-MDMs; and Section 5 concludes this paper.

\section{Preliminaries}
\subsection{(T2T-type) Inter-MDM}
\begin{figure}[tb!p]
  \begin{center}			
  \includegraphics[width=\linewidth]{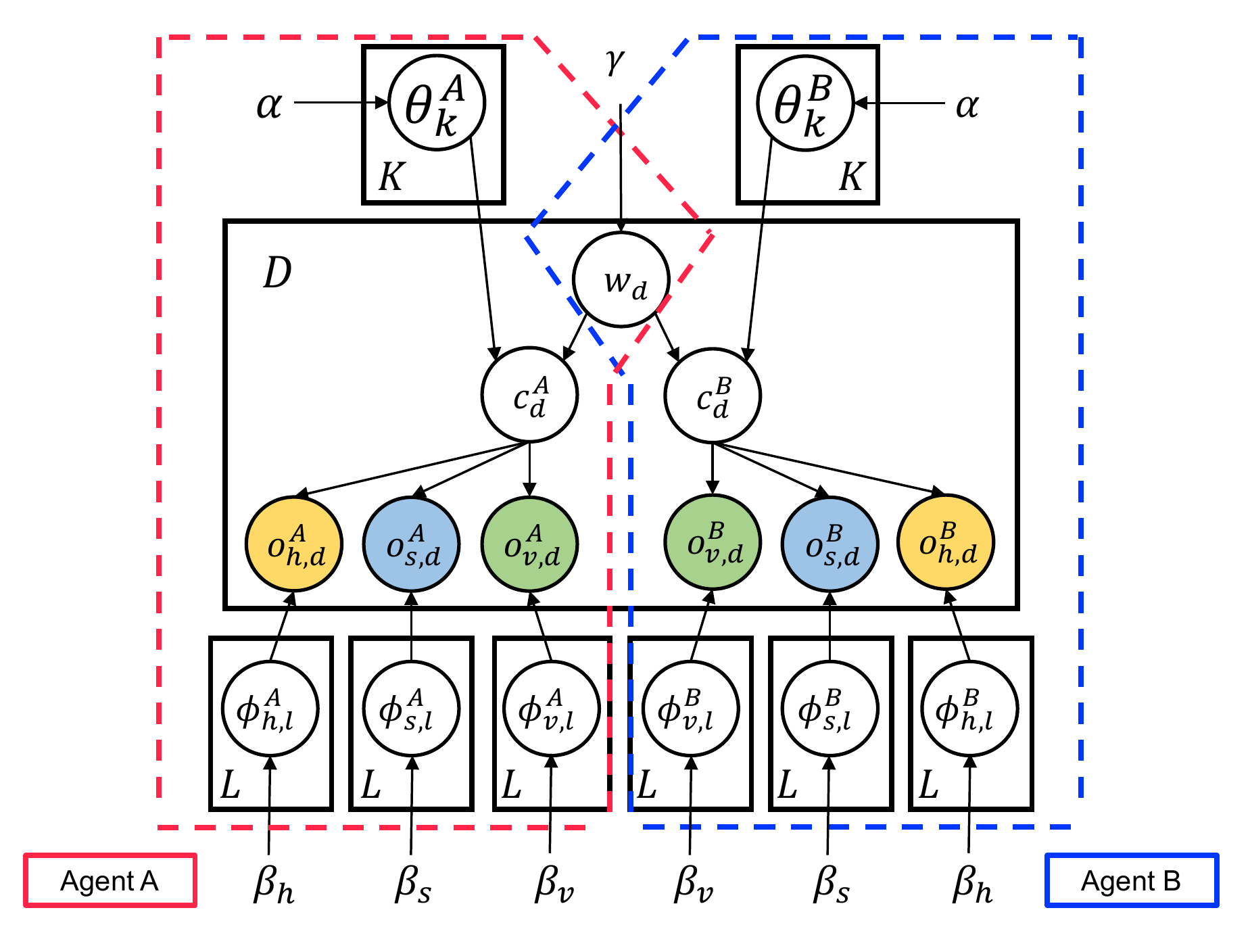}
  \caption{Probabilistic graphical model of the T2T-type Inter-MDM\cite{Furukawa20}}
  \label{fig:t2t_gm}
  \end{center}
\end{figure}

The section introduces the original Inter-MDM, that is, the T2T-type Inter-MDM.
The Inter-MDM is proposed to model the symbol emergence between two agents~\cite{Furukawa20}.
It is based on a model proposed by Hagiwara et al.~\cite{Hagiwara19}, in which each agent forms categories from the information of a single modality.
Figure~\ref{fig:t2t_gm} shows the probabilistic graphical model of the T2T-type Inter-MDM.
The PGM combines two MDMs using a shared prior $w$. 
The MDM is a reduced version of MLDA, which is used in many cognitive and developmental robotics studies to enable a robot to perform object categorization in the Bayesian approach.

Each agent observes multimodal sensory information $o^*_{v,d}, o^*_{s,d}$, and $o^*_{h,d}$. Superscript $*$ corresponds to A or B (i.e., agent index). Subscripts $v, s$, and $h$ represent visual, sound, and haptic information, respectively. $d$ is a target object index. The internal latent variable $c^*_d$ is a category index of the $d$-th object internally estimated by agent $*$. The latent word $w_d$ is a categorical parameter of the prior distribution per $c^*_d$. The global parameters $\theta^*_l, \phi^*_{v,l}, \phi^*_{s,l}$, and $\phi^*_{h,l}$ are global variables for each variable. The others ($\alpha, \beta_v, \beta_s$, $\beta_h$, and $\gamma$) are hyperparameters.  
Please refer to the original paper for further details \cite{Furukawa20}.

\subsection{MH-based naming game for the T2T-type Inter-MDM}
A naming game can be derived as a decentralized inference process of $w$. 
In naming games, each agent tries to categorize the target objects exchanging signs (i.e., names of objects) with other agents.
The strategy for an agent to update their belief about objects and names depends on specific naming game models. 

The abstract of the MH-based naming game for the T2T-type Inter-MDM is elaborated below:
\begin{enumerate}
    \item Each agent performs multimodal categorization using observations $o^*_{v,d}, o^*_{s,d}$, and $o^*_{h,d}$ under the influence of a prior $w_d$.
    \item A speaker agent $Sp \in \{ A, B\}$ samples $w_d \sim P(w_d|c^{Sp}_d)$. This process corresponds to the name utterance by the speaker agent.
    \item A listener agent $Li \in \{ A, B\} \backslash \{ Sp\}$ determines whether or not it will accept the name $w_d$ in a probabilistic manner. The acceptance ratio is
    $\frac{P(c_d^{Li}|{\bf{\Theta}}^{Li},w_d^{Sp})}
{P(c_d^{Li}|{\bf{\Theta}}^{Li},w_d^{Li})}$.
    \item Repeat 1 to 3.
\end{enumerate}

This is called the MH-based naming game, which itself is an MH method for the PGM.
Consequently, $w_d$ is guaranteed to be a sample drawn from $P(w^d| o^A_{v,d}, o^A_{s,d}$, $o^A_{h,d}, o^A_{v,d}, o^A_{s,d}$, $o^A_{h,d})$.
In short, each agent forms categories to share the same category index as $w_d$ and improve the predictability of the multiagent multimodal sensory observations.
Please refer to the original paper for more details.

\subsection{Composition with shared variable and decomposition with SERKET}
The idea of decomposing a large PGM into several modules and inferring latent variables using intra-module inference and inter-module communication is based on (Neuro-)SERKET~\cite{Serket,taniguchi2020neuro}.
Neuro-SERKET is a method that divides the entire model into modules by focusing on the variables shared by the two modules. The total system can perform probabilistic Bayesian inference by performing inter-module updates of the shared node and intra-module updates. SERKET also helps us develop a more complex model by combining several existing models, such as LDA, MLDA, VAE, and GMM\cite{Taniguchi20}. 
The Inter-MDM was developed by reinterpreting the MH-based communication in SERKET as a semiotic communication between two agents.

\begin{figure}[tb!p]
  \begin{center}			
  \includegraphics[width=\linewidth]{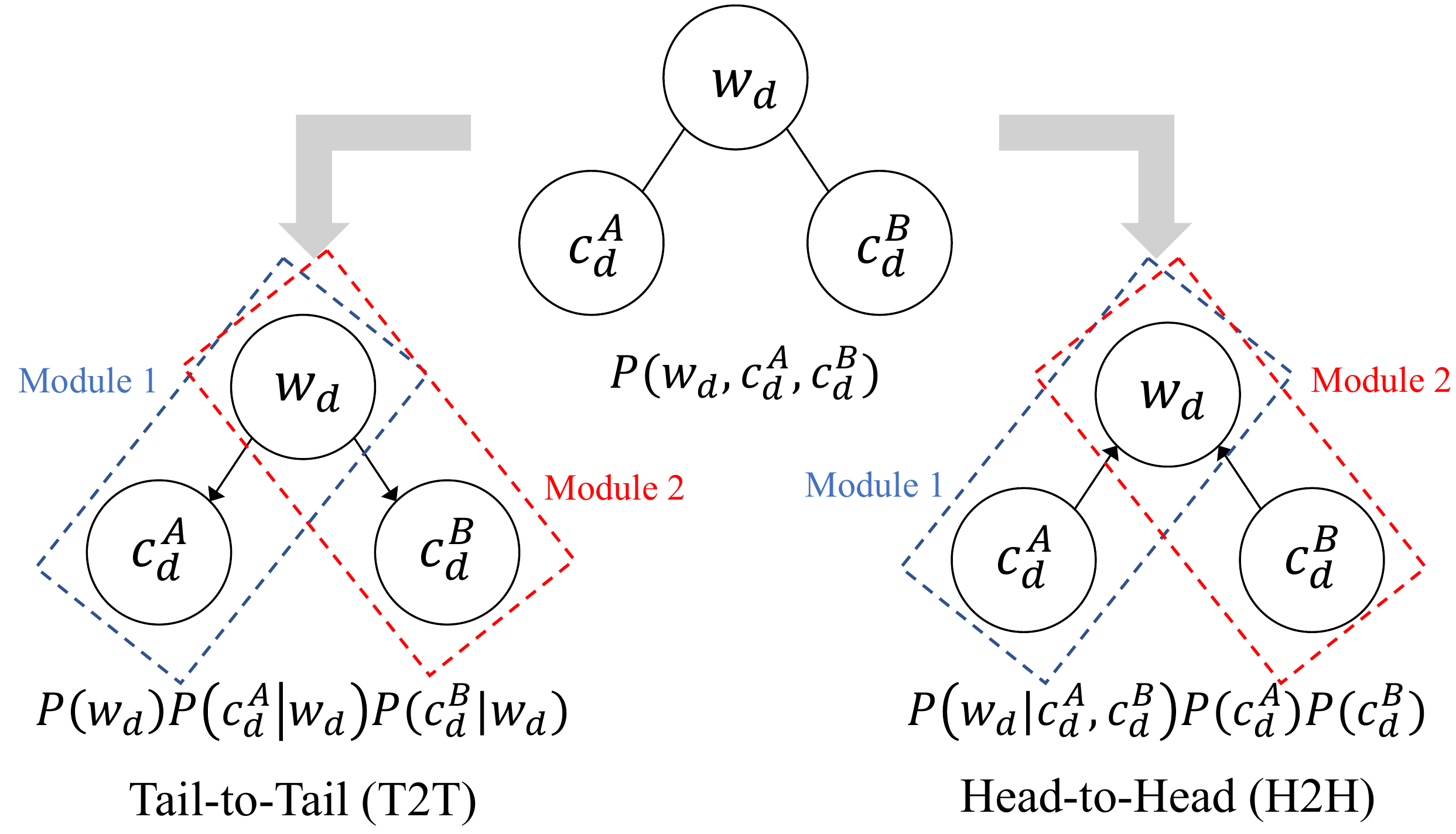}
  \caption{Two ways of introducing symmetric dependencies into two variables sharing a dependent variable (i.e., H2H and T2T) }
  \label{fig:connections}
  \end{center}
\end{figure}

Considering the internal variables $c_d^A,c_d^B$ of the two agents and a shared word variable $w_d$, two different symmetric ways can be implemented to introduce dependencies into the joint distribution $P(w_d,c_d^A,c_d^B)$ from the viewpoint of PGMs as shown in Figure~\ref{fig:connections}.
\begin{description}
\item[T2T connection] \begin{equation}
    P(w_d,c_d^A,c_d^B)=P(w_d)P(c_d^A|w_d)P(c_d^B|w_d)
\end{equation}
\item[H2H connection] \begin{equation}
    P(w_d,c_d^A,c_d^B)=P(w_d|c_d^A,c_d^B)P(c_d^A)P(c_d^B)
\end{equation} 
\end{description}
Figure~\ref{fig:connections} illustrates the two types of dependencies as probabilistic graphical models. 

The H2H connection treatment in Neural-SERKET is presented below. Let $w_d,c_d^A$ denote Module 1 and $w_d,c_d^B$ denote Module 2. 
The joint distribution of $w_d,c_d^A,c_d^B$ can be decomposed into $w_d,c_d^A$-only and $w_d,c_d^B$-only terms by introducing the product-of-expert-like approximation or definition of a generative model~\cite{PoE,Taniguchi20}.
\begin{eqnarray}
P(w_d,c_d^A,c_d^B)&=&P(w_d|c_d^A,c_d^B)P(c_d^A)P(c_d^B) \nonumber \\
&\approx\propto&P_m(w_d|c_d^A)P_m(w_d|c_d^B)P(c_d^A)P(c_d^B) \nonumber \\
&=&P(w_d,c_d^A)\ P(w_d,c_d^B) \nonumber
\end{eqnarray}
where, $P_m(w_d|c_d^*)$ denotes a conditional distribution on $w_d$ in the $*$th module. 
Consequently, the original joint distribution can be decomposed into two parts:
\begin{eqnarray}
P(w_d,c_d^A,c_d^B)&\Rightarrow&P(w_d,c_d^A) \otimes P(w_d,c_d^B) \nonumber
\end{eqnarray}
where, $\otimes$ represents a module composition in the SERKET framework.

\section{H2H-type Inter-MDM}
\subsection{Generative model}

\begin{figure}[tb!p]
  \begin{center}			
  \includegraphics[width=\linewidth]{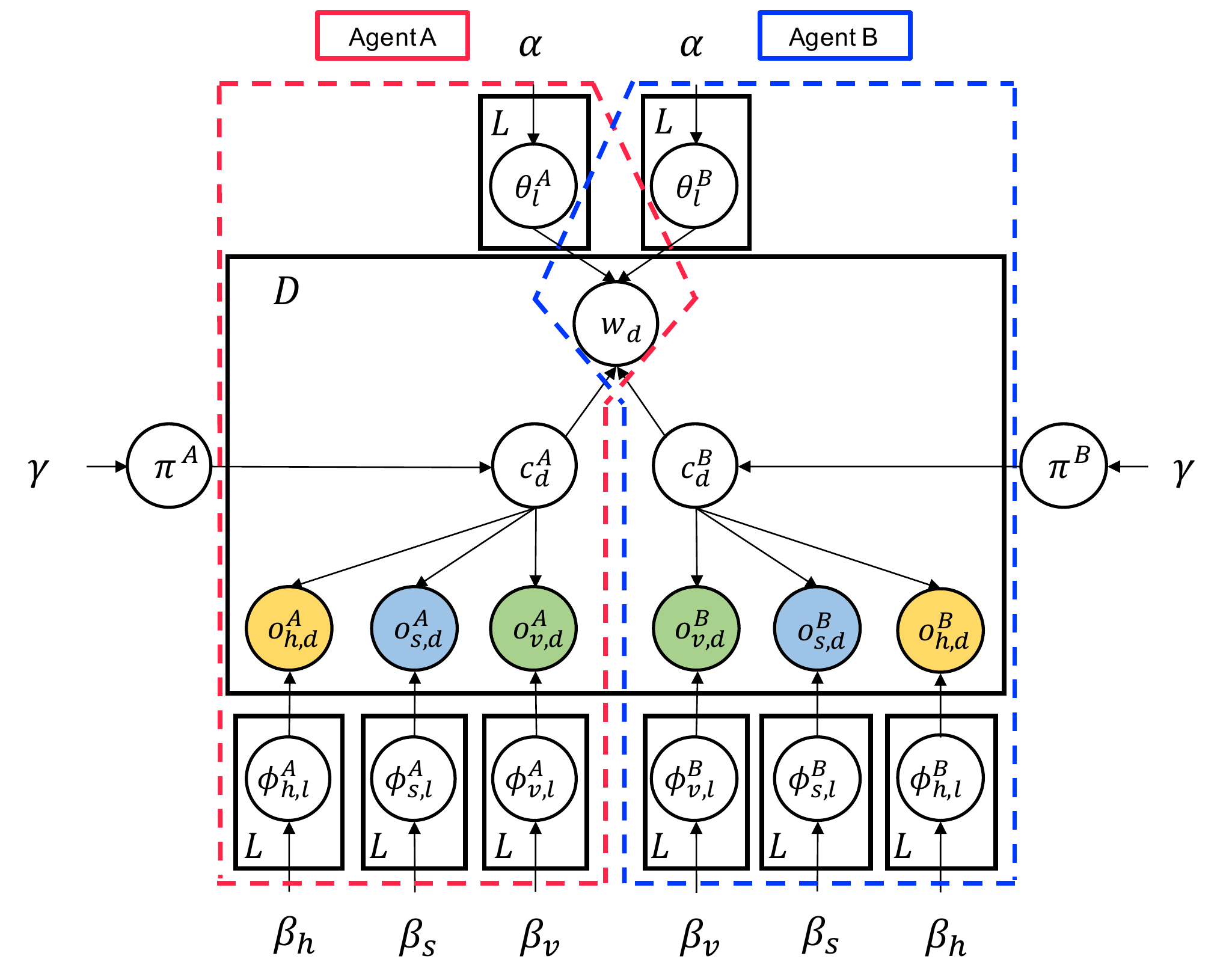}
  \caption{Probabilistic graphical model of the H2H-type Inter-MDM}
  \label{fig:h2h_gm}
  \end{center}
\end{figure}

Figure~\ref{fig:h2h_gm} shows the graphical model of the H2H-type Inter-MDM. We modified the connection across $w_d$ in the T2T-type Inter-MDM into an H2H. The variables and the distributions around the shared variable $w_d$ were consequently changed to keep $c^*_d$ and $w_d$ as discrete variables in the same way as the original T2T-type Inter-MDM. Parameters $\Theta_l^A$ and $\Theta_l^B$ are variables for generating sign $w_d$. Parameter $\pi^*$ is a variable for generating the internal category index $c_d^*$.

The left and right parts in Figure~\ref{fig:h2h_gm} (dashed lines) correspond to the agent A and B models, respectively. Sign $w_{d}$ is the latent word variable connecting agents A and B. $o_{*,d}^A$ and $o_{*,d}^B$ are observations. $c_{d}^A$ and $c_{d}^B$ are category indices. $\phi_{*,l}^A$ and $\phi_{*,l}^B$ are the observation parameters $o_{*,d}^A$ and $o_{*,d}^B$, respectively. $\pi^A$ and $\pi^B$ are the parameters of categories $c_{d}^A$ and $c_{d}^B$, respectively. $\theta_{l}^A$ and $\theta_{l}^B$ denote the probability of sign occurrence for each index. $v$, $s$, and $h$ represent different modalities. $D$ is the number of data. $L$ is the number of categories. $\alpha$, $\beta_*$, and $\gamma$ are hyperparameters.

The process of generating an H2H-type model is described.
Parameters $\phi_{v,l}^A$ $\phi_{s,l}^A$ $\phi_{h,l}^A$ and $\phi_{v,l}^B$ $\phi_{s,l}^B$
$\phi_{h,l}^B$ represent the parameter of the multinomial distributions, which generates features for each observation in the plate ($l\in L$). $\Dir(\cdot)$ represents a Dirichlet distribution.

\begin{eqnarray}
\label{eq:phi_A_h2h}
\phi_{*,l}^A&\sim& {\rm Dir}(\beta_{*})\\
\label{eq:phi_B_h2h}
\phi_{*,l}^B&\sim& {\rm Dir}(\beta_{*})\\
\label{eq:theta_A_h2h}
\theta_{l}^A&\sim& {\rm Dir}(\alpha)\\
\label{eq:theta_B_h2h}
\theta_{l}^B&\sim& {\rm Dir}(\alpha)\\
\label{eq:pi_A_h2h}
\pi^A&\sim& {\rm Dir}(\gamma)\\
\label{eq:pi_B_h2h}
\pi^B&\sim& {\rm Dir}(\gamma)
\end{eqnarray}

For the $d$th object, the following variables are drawn for each $(d\in 1,2,...,D)$. Here, $\Mult(\cdot)$ represents a multinomial distribution, while $\Cat(\cdot)$ represents a categorical distribution. 

    \begin{eqnarray}
    \label{eq:c_A_h2h}
    c_{d}^A&\sim& {\rm Cat}(\pi^A) \\ 
    \label{eq:c_B_h2h}
    c_{d}^B&\sim& {\rm Cat}(\pi^B) \\
    \label{eq:o_A_h2h}
    o_{*,d}^A&\sim& {\rm Mult}(\phi_{*,c_{d}^A}^A)\\ 
    \label{eq:o_B_h2h}
    o_{*,d}^B&\sim& {\rm Mult}(\phi_{*,c_{d}^B}^B)\\ 
    \label{eq:w_d_h2h}
    w_{d}&\sim& \frac {{\rm Cat}(\theta_{c_{d}^A}^A){\rm Cat}(\theta_{c_{d}^B}^B)}{\sum_{w_{d}} {\rm Cat}(\theta_{c_{d}^A}^A){\rm Cat}(\theta_{c_{d}^B}^B)}
    \end{eqnarray}
A product of multinomial distribution is used for the probability distribution on $w_d$.

\subsection{Inference as the MH-based naming game}
\begin{figure}[tb!p]
  \begin{center}			
  \includegraphics[width=\linewidth]{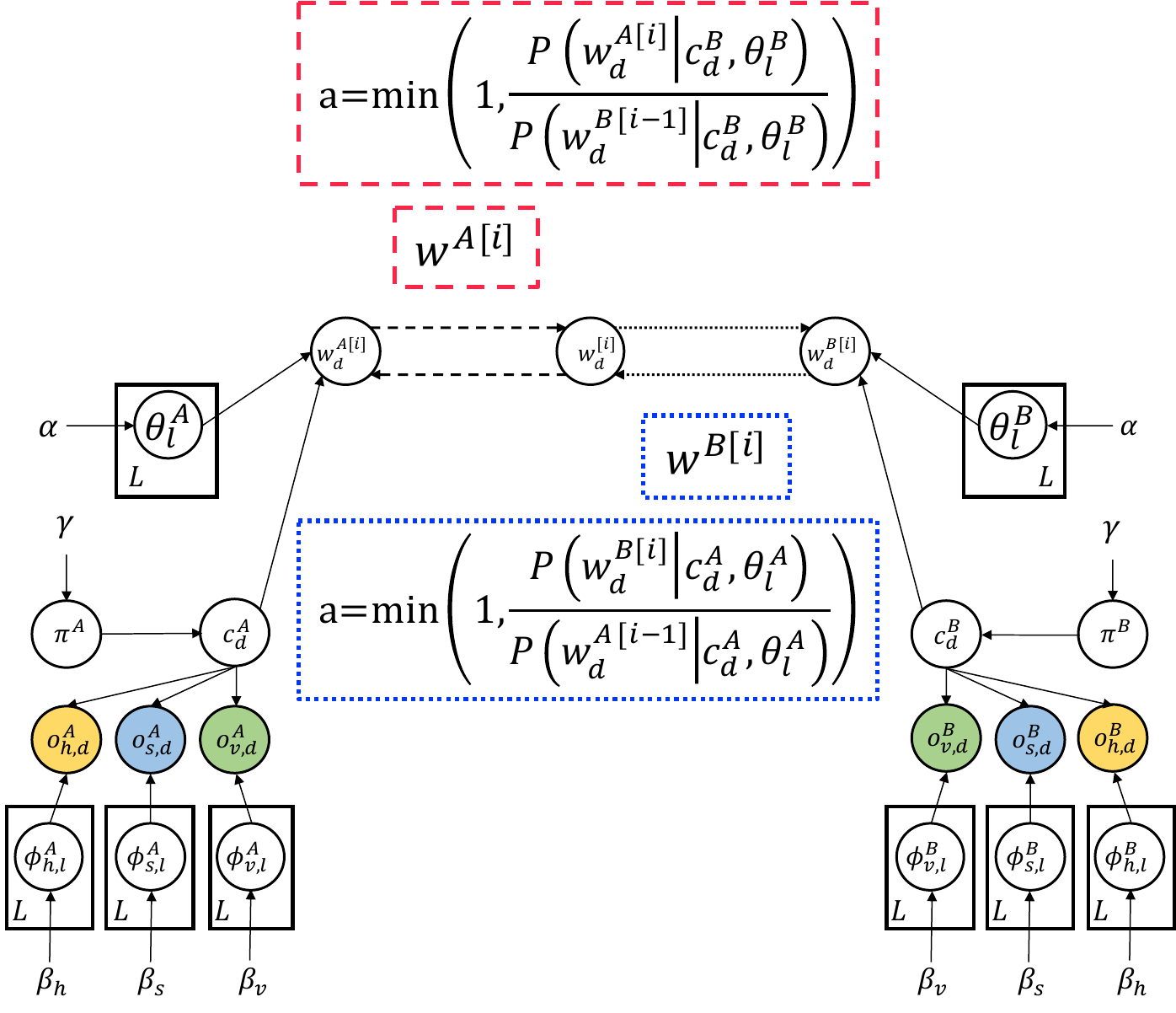}
  \caption{Overview of the inference procedure of the H2H-type Inter-MDM as the MH-based naming game}
  \label{fig:h2h_mh}
  \end{center}
\end{figure}

We will now explain the variable inference following the MH-based inference strategy.
The Neuro-SERKET performs approximate inference by alternately sampling the shared variables (e.g., $w_d$) in the H2H-type Inter-MDM and internal variables.
When the MH-based inference strategy is adopted, the shared variable $w_d$ is sampled using the MH method. A proposal probability distribution is calculated by one module. The other module determines whether or not to accept or reject the value sampled from one module according to the acceptance rate calculated using the internal variables of the module that receives the sign.
This process can be regarded as a naming game. 

The definition of the acceptance rate for the Metropolis--Hastings method is presented as follows:
\begin{eqnarray}
A(z,z^*)&=&{\rm{min}}(1,a) \\
a&=&\frac{P(z^*)Q(z|z^*)}{P(z)Q(z^*|z)}
\end{eqnarray}
where, $z^*$ represents a new sample; $z$ represents a current sample; $P(\cdot)$ represents the target distribution; and $Q(\cdot)$ represents the proposed distribution. The proposed distribution generates samples before accept or reject decision. In the case of the H2H, the target distribution is $P(w_d|c_d^A,c_d^B)$, while the proposed distribution is $P_m(w_d|c_d^A)$ or $P_m(w_d|c_d^B)$.

The acceptance rate in the H2H-type Inter-MDM becomes

\begin{eqnarray}
a&=&\frac{P(w_d^{[i]}|c_d^A,c_d^B)P(w_d^{[i-1]}|c_d^A)}{P(w_d^{[i-1]}|c_d^A,c_d^B)P(w_d^{[i]}|c_d^A)}
\end{eqnarray}
Where, $i$ refers to the index of iterations.
The equation can be transformed using the PoE-based definition of $P(w_d|c_d^A,c_d^B)$.

Figure~\ref{fig:h2h_mh} shows how $w_d$ was inferred using the Metropolis--Hastings method in the H2H-type Inter-MDM. Figure~\ref{fig:h2h_mh} illustrates the partition of agent A and B parts in Figure~\ref{fig:h2h_gm} as distributed modules. The shared variable $w_d$ was alternately sampled from agents A and B to perform approximate inference. In doing so, the other agent decides whether to accept or reject the $w_d$ sampled from one agent according to its acceptance rate. Let $a$ be the acceptance rate according to the definition formula of the acceptance rate of the Metropolis--Hastings method. The acceptance rate $a$ for sampling from agents A to B (sampling from the left side to the right side in Figure~\ref{fig:h2h_mh}) is obtained as follows:

{\footnotesize
\begin{eqnarray}
a&=&\frac{P(w_d^{A[i]}|c_d^A,c_d^B,{\rm \bf{\Theta}}^A,{\rm \bf{\Theta}}^B)P(w_d^{B[i-1]}|c_d^A,{\rm \bf{\Theta}}^A)}
{P(w_d^{B[i-1]}|c_d^A,c_d^B,{\rm \bf{\Theta}}^A,{\rm \bf{\Theta}}^B)P(w_d^{A[i]}|c_d^A,{\rm \bf{\Theta}}^A)} \nonumber \\
&\approx\propto&
\frac{P(w_d^{A[i]}|c_d^A,{\rm \bf{\Theta}}^A)P(w_d^{A[i]}|c_d^B,{\rm \bf{\Theta}}^B)P(w_d^{B[i-1]}|c_d^A,{\rm \bf{\Theta}}^A)}
{P(w_d^{B[i-1]}|c_d^A,{\rm \bf{\Theta}}^A)P(w_d^{B[i-1]}|c_d^B,{\rm \bf{\Theta}}^B)P(w_d^{A[i]}|c_d^A,{\rm \bf{\Theta}}^A)}
\nonumber \\
&=&\frac{P(w_d^{A[i]}|c_d^B,{\rm \bf{\Theta}}^B)}{P(w_d^{B[i-1]}|c_d^B,{\rm \bf{\Theta}}^B)} \nonumber
\end{eqnarray}
}
where, the new sample is $w_d^{A[i]}$; the current (i.e., previously accepted) sample is $w_d^{B[i-1]}$; and $i$ denotes the index of iterations.

\section{Experiment}\label{sec:4}
This study aimed to compare the performances of conventional T2T-type Inter-MDM and the proposed H2H model to determine if the latter can work equivalently with the former in multiagent multimodal categorization. The object categorization accuracy and whether or not the signs (i.e., words) representing object categories are shared among agents should be verified to clarify if the H2H-type model can work in the same manner as the T2T-type Inter-MDM.

We conducted an experiment using the synthetic data used in the original work of the T2T-type Inter-MDM to verify whether or not the H2H-type Inter-MDM can work in the same manner as the original T2T-type Inter-MDM~\cite{Furukawa20}. 

\subsection{Conditions}
The two model types were compared in the four conditions summarized in Table~\ref{tbl:conditions}.
All conditions were used to evaluate the T2T-type Inter-MDM in the original work~\cite{Furukawa20}.
We conducted a comparative experiment in almost the same conditions for the T2T- and H2H-type models despite the PGM differences.

To verify if they work in a similar way, we must determine if the two models can improve the agents' object categorization performance in a collaborative manner through communication and share a symbol system to represent the same object with the same sign.

We used two agents, namely agents A and B, and assumed that each agent had three modalities corresponding to visual, auditory, and haptic sensor information, which are often used in multimodal categorization by a robot\footnote{In this simulation experiment, the labels, vision, sound, and haptics, are labels representing three different observation types. We did not use real sensor information. However, the experiments using real sensor information can be conducted in the same manner as those in \cite{Furukawa20}. In this study, we focused on synthetic data because we aimed to empirically compare the characteristics of the two models. The applicability of the Inter-MDM to the real data was already verified in a previous work. }.  

In Condition 1, agents A and B both had all three modalities. 
We verified if H2H-type Inter-MDM enables agents to share signs with sufficient sensory information for categorization. The Kappa coefficient was adopted to evaluate the sign sharing between the two agents~\cite{Cohen60}.

The Kappa coefficient ($\kappa$) represents the degree of agreement\footnote{The $\kappa$ values denote the following: ($0.81$--$1.00$) almost perfect agreement; ($0.61$--$0.80$) substantial agreement; ($0.41$--$0.60$) moderate agreement; ($0.21$--$0.40$) fair agreement; ($0.00$--$0.20$) slight agreement; and ($\kappa < 0.0$): no agreement~\cite{Criteria}.}.  
The $\kappa$ was calculated as follows:
\begin{eqnarray}
\label{eq:kp}
\kappa &=& \frac{C_o - C_e}{1 - C_e},
\end{eqnarray}
where, $C_o$ is the coincidence rate of the signs between the agents, and $C_e$ is the coincidence rate of the signs between the agents by random chance. 

In conditions 2--4, the agents did not have full observations. 
In conditions 2 and 3, Agent B missed one and two modalities, respectively.
In Condition 4, Agent A also missed one modality.
These conditions made it difficult for agents to perform categorization by themselves without additional information (i.e., communication). 

We adopted the adjusted Rand index (ARI)~\cite{Hubert85} as an evaluation criterion of categorization to quantify the object categorization performance.  
The ARI is a general evaluation criterion widely used in clustering tasks.
\begin{eqnarray}
\label{eq:ari}
\rm ARI &=& \frac{{\rm RI-Expected\:RI}}{{\rm \max (RI)-Expected\:RI}},
\end{eqnarray}
where, RI stands for Rand index.
ARI is the RI adjusted to make ARI$=1$ and ARI$=0$ represent being totally matched and at chance level (i.e., almost at random), respectively. 

We evaluated the categorization accuracy and sign sharing in the original T2T-type. We then proposed the H2H-type Inter-MDM by comparing three different methods. The first method is the communication method proposed for the original Inter-MDM, called the {\it MH-based naming game}. The other methods are the baseline and topline methods.
Two agents independently performed multimodal categorization as the baseline was separated. This was realized by making a listener agent reject all the signs proposed by the speaker agent. We call this method as {\it all rejection}.  
As a topline, the {\it Gibbs sampling} algorithm was used to infer $w_d$.
The Gibbs sampling is a special case of the Metropolis--Hastings method~\cite{bishop2006pattern}. The Gibbs sampling algorithm for the H2H-type Inter-MDM can be derived in the same way as that for the T2T-type Inter-MDM~\cite{Furukawa20}.
The sampling process of $w_d$ required the algorithm to simultaneously use $c^A_d$ and $c^B_d$ (i.e., internal variables/representations of two agents). 
Notably, in the human society, a person can access only his/her internal variables.  
Therefore, nobody can simultaneously use the internal variables of two different agents. Gibbs sampling was computationally ideal, but realistically implausible as a constructive model of a symbol emergence system.  
Gibbs sampling was regarded as a topline method herein.

The synthetic dataset comprised a set of multimodal sensory observations for 15 types of pseudo-objects. 
The observations for the three modalities corresponding to vision, sound, and haptics were represented by a $20$-dimensional feature histogram, that is, the Bag-of-Features representations and $20$ data points were drawn for each modality.
The probabilistic model hyperparameters $\alpha, \beta_*$, and $\gamma$ corresponded to $0.01, 0.001$, and $0.01$, respectively. These parameters were used for the H2H- and T2T-type Inter-MDM and for synthetic data generation.

The number of observed objects $D$ was $150$.
The parameters of $15$ objects were generated from the prior distribution. Accordingly, $10$ observations were sampled for each object. 
The numbers of categories and signs were set to $K = 15$ and $L = 15$, respectively.

\subsection{Experimental results}
Tables \ref{tbl:cond_1}--\ref{tbl:cond_4} present the experiment results. 
The means and the standard deviations of the ARI for each agent and the means of the Kappa coefficients ($\kappa$) in 10 trials with 300 iterations are presented for each condition. 

Table~\ref{tbl:cond_1} shows that the two agents could perform a multimodal object categorization using the three-modality information in all conditions. The Kappa coefficients depicted that the MH-based naming game using the H2H-type Inter-MDM successfully enabled the agents to share sign meanings in a manner similar to that of the T2T type. The performance was even better than the T2T-type model.

Based on Tables~\ref{tbl:cond_2},~\ref{tbl:cond_3}, and \ref{tbl:cond_4}, the two agents in the MH-based naming game and Gibbs sampling could perform multimodal object categorization better than in the all rejection conditions (i.e., without communication).
The categorization performance improvements in the T2T- and H2H-type models were almost the same and even equivalent to those of Gibbs sampling, which simultaneously used the internal information of two agents to perform a categorization.
The Kappa coefficients showed that the MH-based naming game using the H2H-type Inter-MDM successfully enabled the agents to share sign meanings in the same manner as the T2T-type model. Its performance was even better than the T2T type. 

In summary, the H2H-type Inter-MDM and its MH-based naming game work in the same manner as the original T2T-type Inter-MDM and its MH-based naming game in terms of sharing signs and multimodal categorization. As regards its sign sharing performance, the H2H-type Inter-MDM even showed a better performance than the T2T-type model.

\begin{table*}[tb!]
  \begin{center}
  \caption{Summary of the agent conditions in the experiment}

    \begin{tabular}{ c c | c c c | c c c} \hline
     & & & Agent A & & & Agent B & \\
    Condition & Description & \shortstack{Sensor 1\\ (Vision)} & \shortstack{Sensor 2\\(Sound)} & \shortstack{Sensor 3\\(Haptics)} & \shortstack{Sensor 1\\ (Vision)} & \shortstack{Sensor 2\\(Sound)} & \shortstack{Sensor 3\\(Haptics)}\\ \hline
    Condition 1  &  Full perception & $\checkmark$ & $\checkmark$ & $\checkmark$ & $\checkmark$ & $\checkmark$ & $\checkmark$\\ 
    Condition 2  & Lack of one modality in an agent & $\checkmark$ & $\checkmark$ & $\checkmark$ & $\checkmark$ & $\checkmark$ & \\
    Condition 3  & Lack of two modalities in an agent & $\checkmark$ & $\checkmark$ & $\checkmark$ &$\checkmark$   &  & \\
    Condition 4  & Lack of one or two modalities in each agent & $\checkmark$ & $\checkmark$ &  & &  &$\checkmark$   \\ \hline
    \end{tabular}\vspace{-4mm}
  \label{tbl:conditions}
  \end{center}
\end{table*}

\begin{table*}[bt!p]
\centering
\caption{Condition 1 results}
\begin{tabular}{c|l|ccc}
\hline
Model & Method & \shortstack{ARI (Agent A)\\ Mean (SD)} &\shortstack{ARI (Agent B)\\ Mean (SD)} & \shortstack{Kappa coefficient\\$\kappa$} \\  \hline\hline
&MH-based naming game                & 0.881 (0.031) & 0.886 (0.035) & 0.947 (0.046) \\  
Tail-to-tail &All rejection (baseline) & 0.883 (0.035) & 0.886 (0.039) & 0.004 (0.019) \\ 
&Gibbs sampling (topline)               & 0.884 (0.033) & 0.886 (0.031) & -- \\  \hline
&MH-based naming game                & 0.881 (0.031) & 0.888 (0.033) & 0.999 (0.003) \\  
Head-to-head &All rejection (baseline) & 0.882 (0.037) & 0.889 (0.037) & 0.004 (0.032) \\ 
&Gibbs sampling (topline)               & 0.881 (0.031) & 0.882 (0.042) & -- \\  \hline
\end{tabular}
\label{tbl:cond_1} \vspace{-2mm}
\end{table*}

\begin{table*}[bt!p]
\centering
\caption{Condition 2 results}
\begin{tabular}{c|l|ccc}
\hline
Model & Method & \shortstack{ARI (Agent A)\\ Mean (SD)} &\shortstack{ARI (Agent B)\\ Mean (SD)} & \shortstack{Kappa coefficient\\$\kappa$} \\  \hline\hline
&MH-based naming game                & 0.888 (0.033) & 0.708 (0.009) & 0.954 (0.024) \\  
Tail-to-tail &All rejection (baseline) & 0.878 (0.037) & 0.650 (0.025) & 0.001 (0.012) \\ 
&Gibbs sampling (topline)               & 0.880 (0.033) & 0.706 (0.009) & -- \\  \hline
&MH-based naming game                & 0.879 (0.033) & 0.704 (0.006) & 0.996 (0.011) \\  
Head-to-head &All rejection (baseline) & 0.885 (0.053) & 0.649 (0.035) & -0.010 (0.022) \\ 
&Gibbs sampling (topline)               & 0.881 (0.047) & 0.705 (0.004) & -- \\  \hline
\end{tabular}
\label{tbl:cond_2}\vspace{-2mm}
\end{table*}

\begin{table*}[bt!p]
\centering
\caption{Condition 3 results}
\begin{tabular}{c|l|ccc}
\hline
Model & Method & \shortstack{ARI (Agent A)\\ Mean (SD)} &\shortstack{ARI (Agent B)\\ Mean (SD)} & \shortstack{Kappa coefficient\\$\kappa$} \\  \hline\hline
&MH-based naming game                & 0.882 (0.055) & 0.453 (0.029) & 0.931 (0.039) \\  
Tail-to-tail &All rejection (baseline) & 0.874 (0.037) & 0.342 (0.019) & -0.011 (0.027) \\ 
&Gibbs sampling (topline)               & 0.880 (0.029) & 0.451 (0.035) & -- \\  \hline
&MH-based naming game                & 0.883 (0.070) & 0.444 (0.016) & 1.000 (0.000) \\  
Head-to-head &All rejection (baseline) & 0.876 (0.031) & 0.348 (0.018) & -0.011 (0.015) \\ 
&Gibbs sampling (topline)               & 0.881 (0.031) & 0.447 (0.020) & -- \\  \hline
\end{tabular}
\label{tbl:cond_3}\vspace{-2mm}
\end{table*}

\begin{table*}[bt!p]
\centering
\caption{Condition 4 results}
\begin{tabular}{c|l|ccc}
\hline
Model & Method & \shortstack{ARI (Agent A)\\ Mean (SD)} &\shortstack{ARI (Agent B)\\ Mean (SD)} & \shortstack{Kappa coefficient\\$\kappa$} \\  \hline\hline
&MH-based naming game                & 0.710 (0.017) & 0.460 (0.042) & 0.943 (0.043) \\  
Tail-to-tail &All rejection (baseline)  & 0.658 (0.027) & 0.348 (0.023) & -0.006 (0.023) \\ 
&Gibbs sampling (topline)               & 0.706 (0.015) & 0.460 (0.014) & -- \\  \hline
&MH-based naming game                & 0.704 (0.010) & 0.450 (0.015) & 0.992 (0.012) \\  
Head-to-head &All rejection (baseline)  & 0.658 (0.024) & 0.352 (0.011) & 0.004 (0.024) \\ 
&Gibbs sampling (topline)               & 0.705 (0.009) & 0.453 (0.023) & -- \\  \hline
\end{tabular}
\label{tbl:cond_4}\vspace{-2mm}
\end{table*}

\section{Conclusion}
We proposed herein the H2H-type Inter-MDM and an MH-based naming game based on the PGM by modifying the original T2T-type Inter-MDM and verifying if it can work in the same manner as the original model. 
The experimental results confirmed that the H2H-type Inter-MDM performance is equivalent to (and sometimes even better than) that of the original T2T-type Inter-MDM. The H2H-type Inter-MDM and its extension can be used to model symbol emergence systems in future studies. Considering the many pre-existing PGMs that take linguistic information as an observation, the T2T to H2H modification is crucial because it means that the symbol emergence idea as a multiagent multimodal categorization can be combined with a wide range of studies on cognitive models using PGMs. 

Explicitly defined hand-crafted feature representations are currently used for symbol emergence. Extending the model to perform symbol emergence on raw image data is our next challenge and future direction. A variational autoencoder (VAE) is a PGM type; hence, a similar model can be developed based on multimodal VAEs. Another challenge is how to deal with multiple words and sentences. The current model assumes that the share variable $w_d$ is a categorical variable. That is, an emerged symbol is limited to words and does not have compositionality. 
By changing the $w_d$ part of the proposed H2H type from a one-hot vector to a Bag-of-Words (BoW), we will be able to extend the former step from words to multiple words, even if it is very difficult to treat BoW as a prior of a probability distribution like in the T2T-type model. Extending the two-agent naming game to a multiagent naming game that involves more than three agents is another challenge.

\bibliographystyle{IEEEtran}
\bibliography{bibliography,cpc}

\end{document}